\pdfoutput=1
\documentclass[11pt]{article}

\usepackage{acl}

\usepackage{times}
\usepackage{latexsym}
\usepackage{enumitem}
\usepackage[explicit]{titlesec}

\titlespacing{\paragraph}{%
  0em}{%              left margin
  0\baselineskip}{% space before (vertical)
  0\baselineskip}%   
\usepackage[T1]{fontenc}
\usepackage[utf8]{inputenc}
\usepackage{graphicx}
\usepackage{colortbl}
\usepackage{graphicx}

\usepackage{microtype}

\definecolor{darkGray}{gray}{.6}
\definecolor{mediumGray}{gray}{.8}
\definecolor{lightGray}{gray}{.94}
\title{\textsc{EntSUM}: A Data Set for Entity-Centric Summarization}

\author{Mounica Maddela$\mathsection$\thanks{* Equal Contribution \newline $\mathsection$Work done during an internship at Bloomberg} \\
  Georgia Institute of Technology \\
  \texttt{mmadela3@cc.gatech.edu} \\\And
    Mayank Kulkarni\footnotemark[1] \\
  Bloomberg \\
  \texttt{mkulkarni24@bloomberg.net} \\\AND
     Daniel Preo\c{t}iuc-Pietro \\
  Bloomberg \\
  \texttt{dpreotiucpie@bloomberg.net}}
\begin{document}
\maketitle
\begin{abstract}
Controllable summarization aims to provide summaries that take into account user-specified aspects and preferences to better assist them with their information need, as opposed to the standard summarization setup which build a single generic summary of a document.
We introduce a human-annotated data set (\textsc{EntSUM}) for controllable summarization with a focus on named entities as the aspects to control.
We conduct an extensive quantitative analysis to motivate the task of entity-centric summarization and show that existing methods for controllable summarization fail to generate entity-centric summaries.
We propose extensions to state-of-the-art summarization approaches that achieve substantially better results on our data set.
Our analysis and results show the challenging nature of this task and of the proposed data set.\footnote{The data set is available at: \url{https://zenodo.org/record/6359875} }\footnote{The code is available at: \url{https://github.com/bloomberg/entsum} }
\end{abstract}

\section{Introduction}

\begin{figure}
  \includegraphics[width=\linewidth]{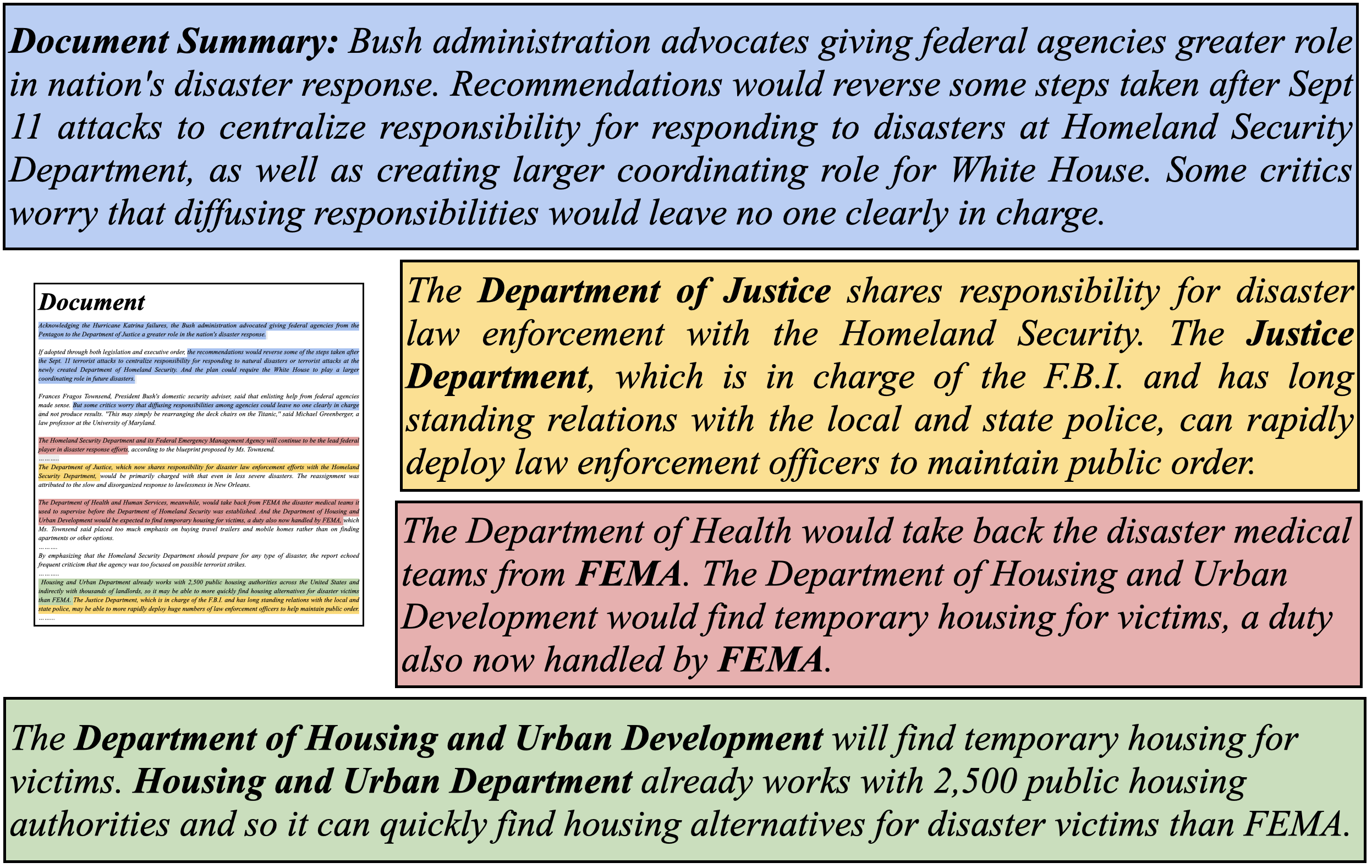} 
  \caption{Example of a generic summary (blue), with three entity-centric summaries from \textsc{EntSUM} focusing on the entities in \textbf{bold}.}
  \label{fig:example}
\end{figure}

Automatic summarization is a core NLP problem that aims to extract key information from a large document and present it to the user with the role of assisting them to digest the core information in the document faster and more easily. However, each user may have a distinct information need and generating a single summary for a document is not suitable for all readers of the document. Recently, various setups for summarization were proposed such that user preferences can be taken into account in the summarization process. These include providing guidance signals such as summary length \cite{kikuchi-etal-2016-controlling}, allowing users to provide terms of interest such as aspects \cite{amplayo-etal-2021-aspect} or entities \cite{fan-etal-2018-controllable} or providing users the flexibility to interact with the summary and explore new facets of interest \cite{avinesh2018sherlock}. The development of such methods may be paramount in enabling the wide-spread usability of summarization technology. Figure \ref{fig:example} shows an example of a document, its generic summary and summaries controlled through salient named entities in the original document.

High quality reference data sets are needed to foster development and facilitate benchmarking. Most summarization data sets are obtained using opportunistic methods such as using abstracts written by editors or librarians when indexing documents. These are by default generic, thus not applicable to controllable summarization. Initial research in this area used small scale human annotations to compare between controllable and generic summarization methods \cite{fan-etal-2018-controllable,ctrlsum}, but these can be prone to biases or qualitative issues, offer only relative quality measurement and do not allow for replicable comparisons between multiple methods or model tuning.

Thus, this paper introduces a new data set for controllable summarization focusing on entities as control aspects given these are usually key aspects in documents and their summaries. The data set consists of 2,788 human-generated entity-centric summaries across 645 documents that are obtained using a strict quality control process mechanism involving several intermediate annotation steps which can be further used in modelling and analyses such as identifying sentences relevant to an entity. The summaries are elicited largely to merge the most important content in a coherent way, while maintaining factuality during the summary creation process.

%The summaries that are elicited are mostly extractive in order to maintain factuality in the summary creation process.

Our data set demonstrates the distinct nature of the entity-centric summarization as opposed to generic summarization and that methods proposed to date for controllable summarization fail at this task. We propose adaptations of state-of-the-art extractive and abstractive summarization methods that significantly improve performance when compared to generic summaries. Our contributions are:
\begin{itemize}[noitemsep,topsep=0pt,leftmargin=1em]
    \item the first annotated data set for controllable summarization with entities as targets for control (\textsc{EntSUM} - Entity SUMmarization);
    \item quantitative data set analysis that highlights the challenges and distinctiveness of this task;
    \item evaluation of generic and also controllable summarization methods on the \textsc{EntSUM} data set;
    \item adaptations of extractive and abstractive summarization methods for performing entity-centric summarization when trained with generic summaries only.
\end{itemize}

\section{Related Work}

\textbf{Controllable summarization} was proposed with the goal of allowing users to define high-level attributes of summaries such as length, source-style or entities \cite{fan-etal-2018-controllable}. Methods relied on adapting existing summarization methods such as CNNs \cite{fan-etal-2018-controllable} or BART \cite{ctrlsum} by pre-pending the controls to the training data and presenting the target control only in inference. However, these methods were only evaluated by comparison to generic summarization methods using human judgments, which can suffer from biases and qualitative issues.

Closely related to controllable summarization, \textbf{guided summarization} also uses an input guidance variable in addition to the document when generating the summary \cite{dou-etal-2021-gsum}. This is different to controllable summarization because the goal of the guidance signal is to generate an improved generic summary by using the guidance to increase faithfulness and quality. Guidance signals explored in past research include summary length \cite{kikuchi-etal-2016-controlling,liu-etal-2018-controlling,sarkhel-etal-2020-interpretable}, keywords \cite{li-etal-2018-guiding,saito2020}, relations \cite{Jin_Wang_Wan_2020} or highlighted sentences \cite{liu2018generating}.

\textbf{Opinion summarization} is the task of automatically generating summaries for a set of reviews about a specific target and usually involves inferring the aspects of interest, predicting sentiment towards them and generating a summary from the extracted sentences \cite{kim2011comprehensive,angelidis-lapata-2018-summarizing}. \citet{amplayo-lapata-2021-informative} studied zero-shot controllability to generate need-specific summaries for movie reviews and evaluated using human comparison judgments.

Contemporaneous to this work, controllable multi-document summarization for aspects in reviews was introduced \cite{angelidis2021,amplayo-etal-2021-aspect}. This work created two data sets used for testing, one focusing on six aspects in hotel reviews (\textsc{SPACE}) and another focusing on 18 aspects for product reviews (\textsc{OPOSUM+}), both obtained using a multi-step annotation process related to the one we use in this paper.

\textbf{Interactive Summarization} is a technique which aims to provide to an interactive faceted summarization of a set of documents and help the user inquire for more information via suggested or free-text queries \cite{avinesh2018sherlock,shapira-etal-2021-extending,hirsch-etal-2021-ifacetsum}. This setup is focused on a multi-document scenario where relevant content to a target concept is retrieved, then fed to a generic abstractive summarization method. 

Recently, \citet{hsu-tan-2021-decision} proposed \textbf{decision-focused summarization}, where the goal is to summarize information across multiple documents with the goal of aiding a human to forecast an outcome.

\section{The \textsc{EntSUM} Data Set}

This section details the collection and annotation process for data set creation. We focus on entities as the aspect to control because named entities are central actors in most news articles and entities are key aspects that make good summaries, together with events and facts. Initial work on controllable summarization considered entities as one of the target for controls \cite{fan-etal-2018-controllable,ctrlsum}.

Most large-scale summarization data sets were obtained opportunistically by mining existing sources of documents and their generic summaries expressed either as titles \cite{narayan-etal-2018-dont}, bullet points \cite{hermann2015teaching} summaries created for indexing purposes \cite{sandhu2008} or TL;DR's created by scientific paper authors \cite{cachola-etal-2020-tldr}. However, we could not identify any similar proxies for entity-centric summaries. Thus, we created the \textsc{EntSUM} data set through a manual annotation process.

\begin{figure*}[t!]
 \includegraphics[width=\linewidth]{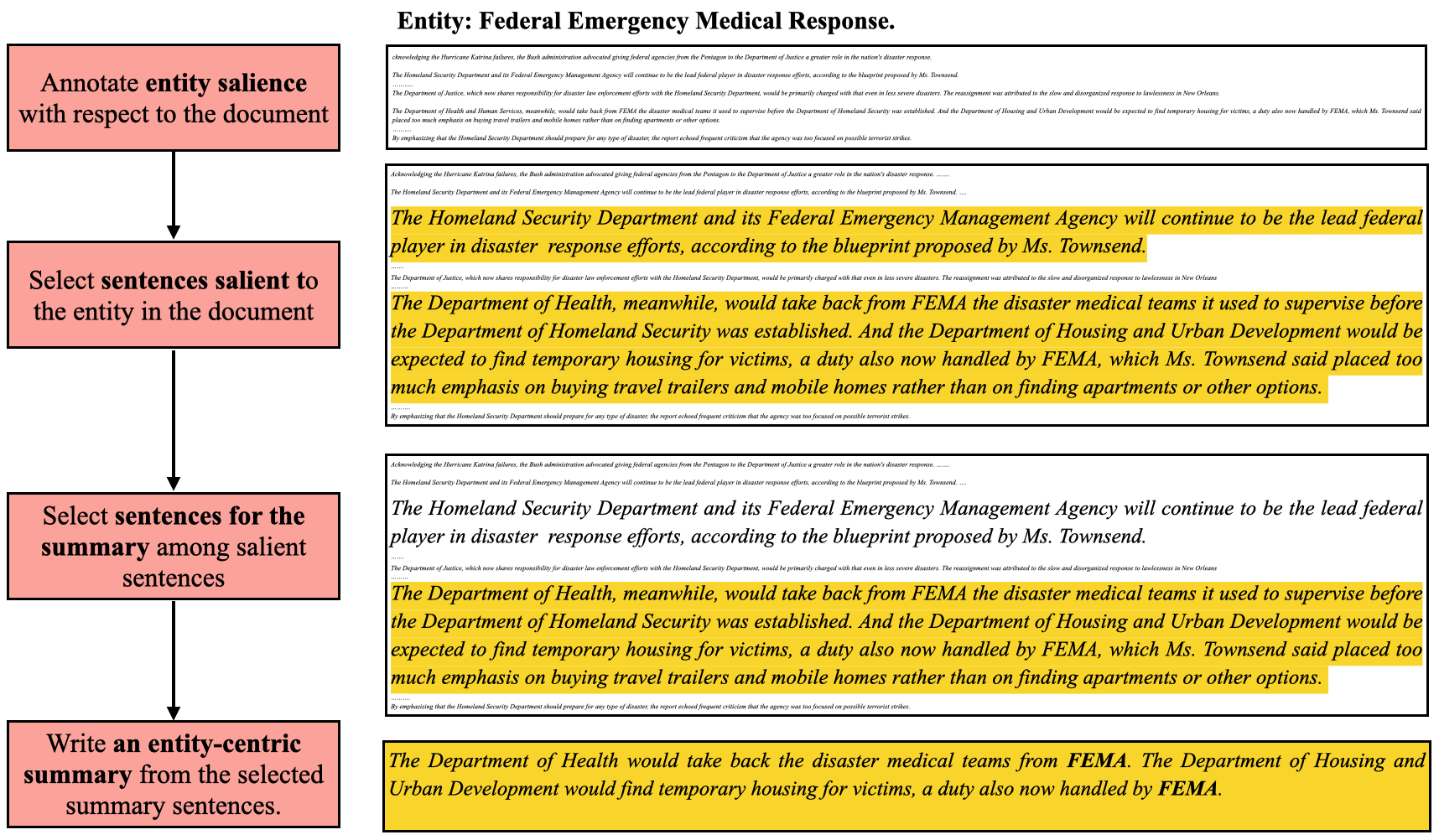} 
 \caption{Annotation pipeline of \textsc{EntSUM}}
 \label{fig:ann_pipeline}
\end{figure*}

\subsection{Task}
Given a document and entity pair, where the entity is a named entity mentioned in the document, the goal of the annotation is to obtain a summary capturing important information about the entity in that document.

\subsection{Data Collection and Preparation}

Our entity-centric summarization data set consists of news articles from the \textit{The New York Times Annotated Corpus} (\textsc{NYT}) \cite{sandhu2008}, which consists of 1.8 million articles written between 1987 and 2007. Around 650k articles in the corpus contain article summaries written by library scientists for indexing purposes. We choose to annotate documents from the NYT data set to enable comparison to generic summaries. We selected the NYT data set instead of other popular summarization data sets (e.g. CNN/DailyMail) because of the clarity of the data licensing terms on the NYT corpus for research purposes \cite{sandhu2008}.

We use the NYT test set as defined in \cite{kedzie-etal-2018-content} to sample the articles used in the \textsc{EntSUM} data set, as we envision the data set will be used primarily for evaluation purposes. %\citet{li-etal-2016-role} partitioned the corpus based on date, with the test set containing articles published in 2006 and 2007. Training and validation data sets contain articles prior to 2005 and from 2005 respectively.
We removed documents with over 1500 words, as we found the majority of these are opinion articles not involving many entities. We split the rest of the documents into sentences and identified named entities using Flair, a high performing system for named entity recognition \cite{akbik2019flair} which identifies Organizations, Person and Location entities. We only select for annotation entities that are Organization and Persons because Locations are usually not salient to the document, thus do not play an active role in the article. From this set, we randomly sampled 10,000 entities spanning 693 documents.

\subsection{Annotation Process}

Summarization is a highly subjective task because the notion of salient information in a document is user-specific and task-dependent \cite{iskender-etal-2020-towards}. There has been relatively little work on the topic of designing annotation guidelines. The most common method to collect summaries is to ask annotators to summarize the document within a specific length limit \cite{harman-over-2004-effects, dangduc}. However, such methods are prone to subjective bias with a low human agreement about the content in the summary \cite{li2021subjective}. Therefore, to ensure quality of the annotation process, we propose a multi-step approach to collect entity-centric summaries that has similarities to the collection method for opinion summarization \cite{angelidis-lapata-2018-summarizing}. Splitting the tasks in multiple steps allows us to ensure quality of the data set through adjudication across multiple annotations at each step which reduces error propagation across tasks. Figure \ref{fig:ann_pipeline} shows an overview of the four-step annotation process.

\subsubsection{Entity Salience}

The first tasks judges if an automatically extracted entity is really a named entity and how salient it is to the source document \cite{gamon2013identifying,gamon2013understanding,dojchinovski-etal-2016-crowdsourced,trani2016sel}. We do this to keep only salient entities for generating summaries, as others are not important targets for entity-centric summaries and may not have enough related content to produce a summary.

Given an article and an entity in the article, we asked the annotators to rate the salience of the entity with respect to the article on a four point scale ranging from not salient (1), through low salience (2), medium salience (3) and high salience (4), similar to \citet{trani2016sel}.
%We highlighted the entity string in the document to improve the document readability for the sub-task.

We collected 2 independent annotations for each entity and increased redundancy up to 5 if there was disagreement. We take the salience rating as the average of all individual ratings. We observe that entities with an average rating $<$ 1.5 are generally mentioned once in the document and, therefore can not have a meaningful summary. We remove these entities, resulting in 3,846 entities. We further grouped the entity mentions from each document using substring matching because multiple entity strings can refer to the same entity (e.g. \textit{Barack Obama} -- \textit{Obama}). After grouping, we obtain 2,788 entities to use in subsequent tasks.

\renewcommand{\arraystretch}{1.1}
\begin{table*}[t!]
    \small
    \centering
    \begin{tabular}{l|c|c|c|c|c}
    \hline
    \rowcolor{mediumGray}\textbf{Metric} & \textbf{Overall} & \multicolumn{2}{c|}{\textbf{Entity Type}} & \multicolumn{2}{c}{\textbf{Entity Salience}} \\
    \hline
    \rowcolor{lightGray}&  & PER & ORG & Medium & High  \\
    \hline
    Number of Salient Entities (\textbf{Task 1}) & 2788 & 1741 & 1047 & 2100 & 688 \\
    Sentences with entity mentions  & 3.95 & 4.21 & 3.46 & 3.36 & 5.65 \\
    Entity Salient Sentences (\textbf{Task 2}) & 5.80 & 6.34 & 5.02 & 4.95 & 8.56 \\
    Entity-Centric Summary Sentences (\textbf{Task 3})  & 2.49 & 2.59 & 2.28 & 2.33 & 2.66 \\
    Summary word length (\textbf{Task 4}) & 81.7 & 84.9 & 76.1 & 78.6 & 88.2 \\
    Summary char length (\textbf{Task 4}) & 444.3 & 458.1 & 421.7 & 432.1 & 482.9 \\
    \hline 
    \end{tabular}
  \caption{Statistics for the output of each task in our entity-centric summary annotation pipeline, overall and across entity types and salience scores as annotated in Task 1. \textbf{PER} and \textbf{ORG} refer to ``Person'' and ``Organization'' entity types respectively.}
  \label{table:ann_stats}
\end{table*}

\renewcommand{\arraystretch}{1.1}
\begin{table*}[t!]
    \small
    \centering
    \begin{tabular}{l|c|p{0.5cm}p{0.5cm}p{0.6cm}|p{0.5cm}p{0.5cm}p{0.6cm}|p{1.0cm}p{0.7cm}|p{0.9cm}p{0.8cm}}
    \hline
    \rowcolor{mediumGray}    & & \multicolumn{3}{c|}{\textbf{Avg. summary len.}} & \multicolumn{3}{c|}{\textbf{Avg. article len.}} & \multicolumn{2}{c|}{\textbf{Compression Ratio}} & \multicolumn{2}{c}{\textbf{\%  novel ngram}} \\
    \rowcolor{lightGray}    \textbf{Data set} & \textbf{Size} & \textbf{sents.} & \textbf{words} & \textbf{char.} & \textbf{sents.} & \textbf{words} & \textbf{char.} & \textbf{article} & \textbf{salient} & \textbf{unigram} & \textbf{bigram} \\
    \hline
    \textsc{NYT} & 41,265 & 4.9 & 117 & 677 &  36.9 & 1021 & 5471 & 0.12 & -- & 11.5 & 39.5\\
    \textsc{CNNDM} & 312,085 & 3.7 & 56 & 297 &  33.1 & 782 & 3998 & 0.089 & -- & 13.3 & 49.95\\
    \textsc{EntSUM} & 2788 & 2.5 & 81 & 444 & 34.4 & 1002 & 5319 & 0.09 & 0.62 & 0.82 & 5.93 \\
    % \textsc{EntSUM$_{ONE}$} & 1921 &  2.4 & 81 & 441 & 34.7 & 1014 & 5379 & 0.093 & 0.64 & 0.61 & 4.95\\
    % \textsc{EntSUM$_{TWO}$} & 867 &  2.6 & 83 & 453 & 33.7 & 977 & 5189 & 0.094 & 0.58 & 1.3 & 8.09\\
    \hline
    \end{tabular}
  \caption{Comparison of the existing document summarization data sets with \textsc{EntSUM}. We report the  corpus size, average article and summary length (in terms of words, sentences, and characters), and percentage of novel n-grams in the summary when compared to the article. We also report the compression ratio of the summary with respect to the original article text and the entity-specific salient text selected by annotators.}
  \label{table:overall_comp_stats}
\end{table*}

\subsubsection{Salient Sentence Extraction}

The second task aims to identify all sentences in the article that are salient to the target entity. To facilitate the process, we displayed all sentences in a document in a tabular format and premarked sentences that contain the given entity mention. The annotators can add additional sentences or remove existing ones. We also asked the annotators to keep the salient sentences as complete as possible by including the sentences that resolve any references in the initially selected sentences.

We collected three annotations for each document and entity pair resulting in three annotations for all sentence and entity pairs. We assigned each sentence a binary label (salient to the entity or not) using majority vote across the three annotations.

Table \ref{table:ann_stats} shows the average number of salient sentences (5.80) is much higher than the average number of premarked sentences (3.95), indicating this task resulted in an expansion from only using the sentences that explicitly mention the target entity.

\subsubsection{Entity-Centric Summary Sentences}

The third task aims to identify the sentences in the article that are used to make up the entity-centric summary. We display the sentences of the document in a tabular format with the salient sentences extracted from the previous task highlighted and allowed the annotators to select only from these sentences. We instructed the annotators to first select up to 3 sentences and add up to 3 more sentences if these are needed to provide context.

\subsubsection{Entity-Centric Summary}

The final task is to write a coherent summary for the entity in the document of up to 150 words using the summary sentences selected previously. This task was performed together with the third task, as they are tightly coupled, to limit cognitive load and to be able to control for quality by comparing selected summary sentences.

As this is a labor intensive task, we collected two annotations for a subset of the target entities (867 out of 2,788) to measure agreement. We provide both summaries in the data set release in order to facilitate evaluation with multiple references. The annotated summary sentences represent only 41.3\% of all salient sentences across all the tasks. Table \ref{table:ann_stats} shows the annotation statistics.

We note the output of each task is released with the \textsc{EntSUM} data set and can be used when training models, for separate tasks or as auxility tasks in a multi-task learning setup.

\subsection{Data Quality}

We devised multiple tasks to accomplish our goal of ensuring quality throughout the annotation process and to make the complex and subjective task of summarization easier for annotators. We adjudicate annotations across multiple annotators to reduce error propagation, wherein if one task has wrong annotations, the subsequent tasks will have the error propagated.

We use our internal annotation platform for obtaining annotations. The annotation was performed using a group of English-speaking vendors who were hired and trained for completing this task through training sessions and performed the task independently from each other. We do not collect any private information from the annotators and do not release the identity of the annotators together with the data. We conducted several training sessions and initial rounds with the annotators, the results of which were discarded, to ensure the annotators are proficient in the task. The training rounds included 100 items for the first two tasks and 50 items for the latter two for all annotators.

We perform multiple annotations for the upstream tasks. For the entity salience task which is a four-way classification task, we elicit 2 annotations for each item and, if these disagree, we increase redundancy to up to 5 annotations if there is no majority (2 annotations -- 6261 items;  3 annotations -- 3318 items; 5 annotations -- 421 items). For the salient sentence extraction task, we elicit 3 annotations for each item and adjudicate annotations at the sentence level using majority vote.

We report inter-annotator agreement for each task. For the 4-way ordinal entity salience task we observe 0.709 interval Krippendorf's Alpha \cite{krippendorff2011computing}, which corresponds to substantial agreement \citep{artstein2008inter}. The annotators agreed on a single annotation 62.6\% of the time. For the salient sentence selection task, we compute inter-annotator agreement using Krippendorf's Alpha between binary sentence-level judgments and obtain a value of 0.744 Krippendorf's Alpha, which again indicates substantial agreement. All three annotators agreed on the same value for 88.4\% of the sentences.

Selecting the summary sentences is a more subjective task, especially given that all sentences are salient to the target entity. Despite this, the inter-annotator agreement is of 0.539 Krippendorf's Alpha, which is considered good agreement.  Finally, in the summary creation task, we compute ROUGE \cite{lin-hovy-2003-automatic} between the summaries and achieve the following values: ROUGE-1 = 71.7; ROUGE-2 = 62.6 and ROUGE-L = 69.0. We release both summaries in our data set where available, as these could be used as multiple references when computing evaluation metrics.

\subsection{Data Analysis}

\paragraph{Summary Statistics}~Table \ref{table:overall_comp_stats} presents summary statistics relevant to summarization data for the newly introduced \textsc{EntSUM} data set, with the commonly used document generic summarization data sets CNN-DailyMail (\textsc{CNNDM}) and \textsc{NYT}. We note that summaries in \textsc{EntSUM} are shorter than their generic counterparts in the \textsc{NYT} corpus, but longer than those in \textsc{CNNDM}, except for the number of sentences, which is expected as the summaries in \textsc{CNNDM} undergo the most compression as demonstrated by the article compression ratio. \textsc{EntSUM} exhibits the lowest percentage of novel unigrams and bigrams, in line with how our annotation was set up to focus on integrating the original content in a coherent summary.
The entity-specific salient text is significantly shorter than the entire document and, as a result, the summary contains the relevant content without requiring dramatic paraphrasing or compression.

\paragraph{Comparison to Generic Summaries}~Our hypothesis is that a new data set for entity-centric summarization is needed as entity-centric summaries do not align well with generic summaries. We compute ROUGE \cite{lin-hovy-2003-automatic} scores between the entity-centric summaries in \textsc{EntSUM} and their corresponding generic summaries in the \textsc{NYT} corpus, with the following values: ROUGE-1 = 26.2, ROUGE-2 = 9.8 and ROUGE-L = 22.9. Low scores show there is low lexical and content overlap between the entity-centric summaries and their corresponding document summaries, demonstrating the distinctiveness of the entity-centric summarization task.

\paragraph{Entity Type and Salience}~Table \ref{table:ann_stats} shows the task-specific statistics of \textsc{EntSUM} by entity type and salience level separately. We note that the data set has more person entities than organizations and, on average, the related content and summaries associated to people is slightly longer. There are significantly more entities with medium salience values when compared to highly salient entities, which are an average slightly more than one for each document. We note that both sentences with entity mentions and salient sentences to the entities are substantially larger in number for highly salient entities, but there is just a small gap for the entity-centric summaries and sentences, which shows that more selection and compression was achieved for these highly salient entities.

\begin{figure}[h!]
  \centering
  \includegraphics[width=\linewidth]{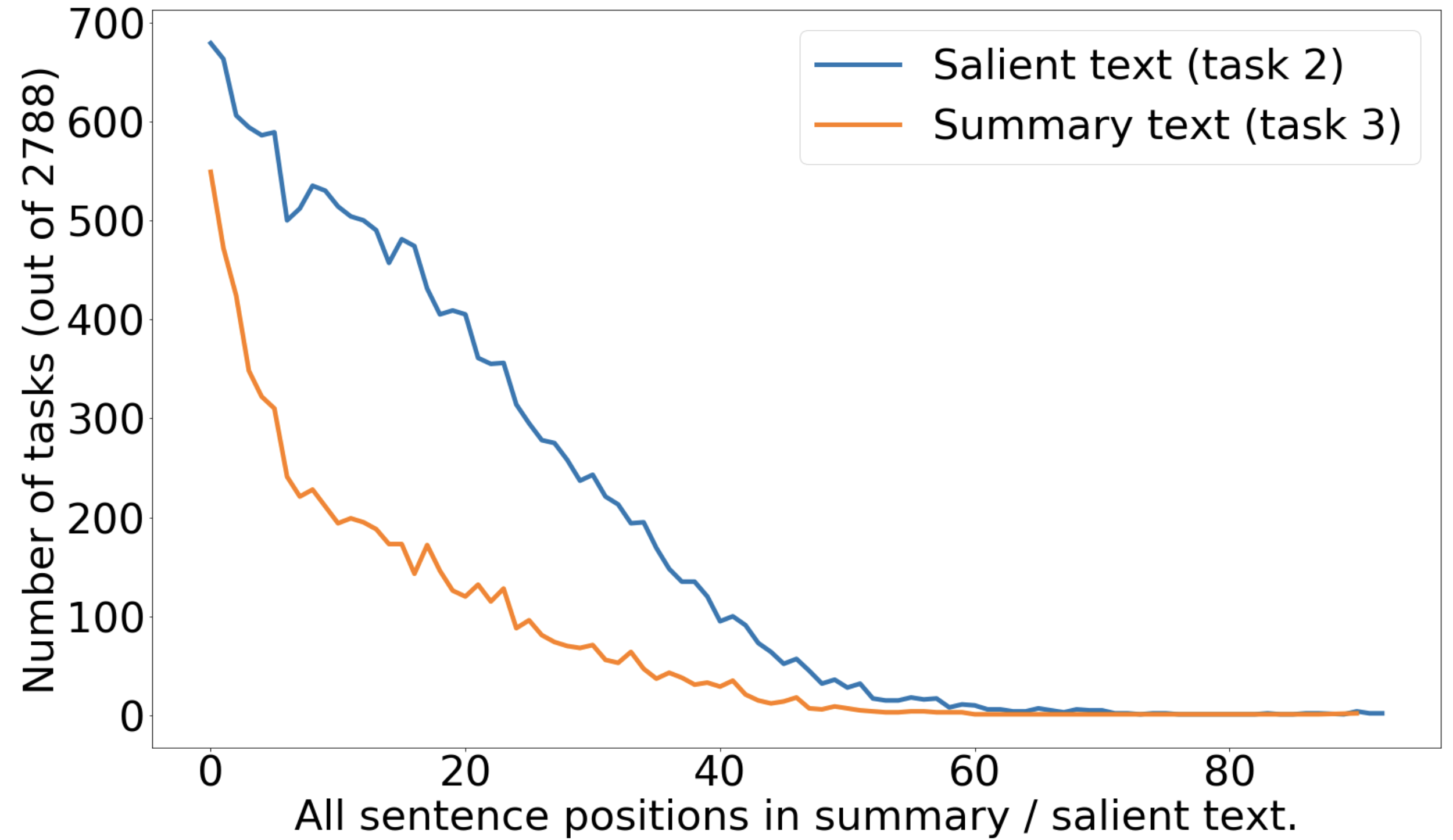}
  \caption{Distribution of sentence positions for salient and summary sentences.}
  \label{fig:pos_dist}
\end{figure}

\paragraph{Sentence Position Distribution}~Figure \ref{fig:pos_dist} shows the position distribution of entity salient and entity-centric summary sentences in the original document. The figure highlights that both types of sentences are more likely to be distributed at the start of the document, which is expected given we are only considering salient entities to the document. We see that sentences used for summaries are even more likely to be towards the start of the document. However, the sentence distribution is not very skewed, with hundreds of summary sentences being present even in position 20 or higher in the original document. This highlights the challenging nature of the data set.

\section{Methods}

For an initial modelling attempt for the \textsc{EntSUM} data set, we evaluate all controllable summarization approaches proposed to date, generic summarization methods, strong heuristics for summarization and a couple of adaptations of state-of-the-art methods for abstractive \cite{dou-etal-2021-gsum} and extractive summarization \cite{liu-lapata-2019-text} to the entity-centric summarization task.

Some of the methods described in this section involve detecting the entity mentions in documents unlabeled with entities in training and/or at inference time. For this, we use a combination of standard methods for NER based on Flair \cite{akbik-etal-2018-contextual} and their coreferent mentions as identified through the SpanBERT coreference system \cite{joshi-etal-2020-spanbert}.

\subsection{Abstractive Methods}

Abstractive summarization uses generation methods to express the content of the original document.

\subsubsection{ConvNet for Controllable Summarization}

We denote through \textbf{ConvNet} the first method for controllable summarization proposed in \citet{fan-etal-2018-controllable}. It adopts a CNN encoder-decoder model for summarization and is trained by replacing entities in the document with placeholders and prepending them to the document. At inference time, only the target entity is prepended to the summary to generate the  entity-centric summary \cite{fan-etal-2018-controllable}. 

\subsubsection{CTRLSum}

\textbf{CTRLSum} \cite{ctrlsum} is a method based on BART \cite{lewis-etal-2020-bart}, a popular Transformer-based sequence-to-sequence model for summarization. CTRLSum is fine-tuned by prepending keywords, in this case all detected entity mentions, to the input document to control the summary \cite{ctrlsum}. At inference time, only the target entity is prepended to the target document to generate the entity-centric summary. 

\subsubsection{GSum}

\textbf{GSum} \cite{dou-etal-2021-gsum} is a document summarization framework that allows for using as input a guidance signal (e.g. keywords, sentences) along with the source document with the goal of improving the generic document summarization task through improving faithfulness. The model architecture consists of a Transformer \cite{NIPS2017_3f5ee243} model initialized with BART \cite{lewis-etal-2020-bart}. The model has two encoders: one to encode the source document and the other to encode the guidance signal. The encoders share the embedding and the encoding layers except for the topmost layer. The decoder first attends to the guidance signal to select the part of the document to focus on and then attends to the document with these guidance-aware representations. The framework allows to include varied guidance signals and demonstrates improvements on generating generic summaries.
%For more details about GSum, please refer to \cite{dou-etal-2021-gsum}. 

\subsubsection{Adapting GSum for Entity-Centric Summarization}

We adapt GSum to generate entity summaries by using the entity information as guidance signal. However, the original GSum implementation used a single generic summary as output for each input document, which is not suitable for our setup in which the output is conditioned on both the input document and the guidance signal (i.e. entity). In addition, we do not have access to gold entity mentions in training and inference and, because we only use \textsc{EntSUM} in evaluation only, we do not have gold reference entity-centric summaries. We create proxies as above for the input and output in training as follows:
\begin{itemize}[noitemsep,topsep=0pt,leftmargin=1em]
    \item for each training and testing (document, entity) pair, we feed the full document and as guidance input either the mention string (\textbf{GSum$_{ent-name}$}) or the sentences that mention the given entity (\textbf{GSum$_{ent-sent}$}) as detected by our NER and coreference approach previously described;
    \item the output summary for each (document, entity) training pair is obtained from the reference entity-agnostic summary as follows: (a) Select at most 3 sentences in the reference that mention the entity; (b) If we obtain less than 3 sentences in the previous step, then select the remaining sentences from the lead 3 sentences that mention the given entity.
\end{itemize}

Note this GSum setup can be used with gold entity mentions, sentences and output if \textsc{EntSUM} data is used in training or development.

\subsection{Extractive Methods}

Extractive summarization methods aim to extract the segments (in this case, sentences) from the original document to form a summary.

\subsubsection{Heuristics}

Selecting the top sentences in a document is a strong heuristic for the document summarization tasks \cite{AAAI1714636}. We evaluate the following variants:

\noindent \paragraph{Lead3$_{ovr}$}~is a generic summarization method that selects the first three sentences in the document irrespective of the target entity.

\noindent \paragraph{Lead3$_{ent}$}~is the entity-aware summarization variant which selects the first three sentences in the document that mention the given entity, as inferred by our NER and coreference resolution approach.

\subsubsection{BERTSum}

\noindent \paragraph{BERTSum}~obtains near state-of-the-art results for extractive summarization \cite{liu-lapata-2019-text}. The method uses the BERT \cite{devlin-etal-2019-bert} encoder to generate representations for each sentence, then models the interactions between these sentences through a BERTSum summarization layer and then predicts the most important sentences from these as the sentences to be part of the generic summary. We evaluate on both all and top 3 predicted sentences to make fair comparisons with Lead3 baselines.

\subsubsection{Adapting BERTSum for Entity-Centric Summarization}

We adapt BERTSum in the training phase by restricting the input only to all the sentences containing the entity string mention and its coreferent mentions, instead of the entire source document. In training, the output entity-centric summary is constructed in a similar way to the GSum training procedure, where we use the generic summary to select top 3 sentences that mention the entity or otherwise up to 3 sentences that mention the entity.

% We have also explored various variations in the BERTSum architecture to support multiple inputs, where one input is the source document and the other is the entity and coreferent sentences or is just the tagged entity string. We attempt to model this through a shared and also independent bi-encoder setup and we explore, self-attention, cross-attention and self-attention in addition to cross-attention in the Summarization layers. We also explore generating entity embeddings through the shared encoder and concatenating these embeddings to each of the source document's sentence embeddings before the summarization layer to provide a stronger entity specific signal. However, we find that all of these experiments result in only in marginal gains over the regular BERTSum architecture.

\subsubsection{Heuristics using Oracle Sentence Information}

Most previous approaches make the realistic assumption that gold entity mentions or other entity-related annotations are not available at inference time. To explore the impact of these, we explore the following additional heuristics:

\noindent \paragraph{Oracle - Lead3$_{ent}$ (salient)}~uses as summary the first three salient sentences selected by annotators during the second step of the annotation pipeline. 

\noindent \paragraph{Oracle - Lead3$_{ent}$ (summary)}~uses as summary the first three sentences selected by annotators for writing the summary. 

We expect these to have high performance given the extractive nature of \textsc{EntSUM} and that these tasks were a prerequisite to writing the summary.

\section{Experimental Setup}

\subsection{Training Data}

We train all non-entity-centric methods on the NYT corpus consisting of  44,382 training and 5,523 validation (document, summary) pairs as specified in \citet{kedzie-etal-2018-content}. However, this data set size increases to 464,339 training and 58,991 validation pairs when training the adapted GSum and BERTSum as each document contains multiple entities resulting in multiple <document, summary> pairs for a single document. 
%To train the entity-centric models, we only use the documents with length $<$ 1500 tokens. \dpp{Why?}.

\subsection{Implementation Details}

\label{ssec:impl_details}

We use the author's implementations for the following methods: CTRLSum,\footnote{\url{https://github.com/salesforce/ctrl-sum}} BERTSum,\footnote{\url{https://github.com/nlpyang/BertSum}} and GSum.\footnote{\url{https://github.com/neulab/guided_summarization}} We reimplement the ConvNet method using the FairSeq library \cite{ott2019fairseq} as described in \citet{fan-etal-2018-controllable}. For all our implementations, we first train on the CNN DailyMail data set and compared to published numbers to ensure we are able to reproduce the original results and then retrain on the NYT data set for reporting our results on \textsc{EntSUM}.

We experiment with various hyperparameter settings for each of the architectures but we find that the original hyperparamters used for training each of the CNN DailyMail models seem to be the most stable and produce the best results.

\subsection{Evaluation}

\renewcommand{\arraystretch}{1.0}
\begin{table*}[t!]
    \small
    \centering
    \begin{tabular}{p{3.6cm}|ccccc}
    \hline
    \rowcolor{mediumGray}       & \textbf{ROUGE-1}  & \textbf{ROUGE-2} & \textbf{ROUGE-L} & \textbf{BERTScore} & \textbf{Avg. Len} \\
    \rowcolor{mediumGray}       & & & & & \textbf{Sent. / Word} \\
    \hline
    \rowcolor{lightGray}    \multicolumn{6}{l}{\textbf{Extractive Summarization Methods}} \\
    \hline
    %  Lead3$_{ovr}$ &  35.4 & 20.2 & 31.9 &  57.6 & 3 / 100 \\ & 32.3 & 16.8 & 28.9 & 59.9   & 3 / 98  \\
     Lead3$_{ovr}$ & 34.44 & 19.14 & 30.97 & 58.32 & 3.0 / 99.38 \\
    %  Lead3$_{ent}$ &  71.3 & 64.3 & 68.2 & 79.4 & 2.7/92 \\ &  62.0 & 52.1 & 58.0 & 81.6   & 2.9 / 93 \\ 
     Lead3$_{ent}$ & \textbf{68.41} & \textbf{60.51} & \textbf{65.03} & \textbf{80.08} & 2.76 / 92.31\\
    %  BERTSum$_{ovr}$  & 34.8 & 19.0 & 31.2 & 57.8     & 3 / 110  \\ & 31.6 & 15.1 & 27.9 & 59.2 & 3.1 / 110  \\
    \hline
     BERTSum$_{ovr}$ & 33.8 & 17.79 & 30.17 & 58.24 & 3.03 / 110.0 \\
     BERTSum$_{ovr-top3}$ & 33.6 & 17.6 & 29.9 & 57.99 & 2.94 / 105.78 \\
     BERTSum$_{ent}$  (Ours) &  65.9 & 58.7 & 62.8 & 77.67  & 4.26 / 128.39  \\
     BERTSum$_{ent-top3}$  (Ours) & \textbf{67.8} & \textbf{59.7} & \textbf{64.4} & \textbf{77.89} & 2.49 / 81.53  \\
    \hline
    \rowcolor{lightGray}    \multicolumn{6}{l}{\textbf{Abstractive Summarization Methods}} \\
    \hline
    %  ConvNet  & 29.6 & 14.3 & 26.6 & 54.1 & 3.9 / 103 \\  & 27.4 & 11.8 & 24.2 & 56.1 & 4.0 / 100 \\
     ConvNet & 28.92 & 13.52 & 25.85 & 54.72 & 3.93 / 102.07 \\
    %  CTRLSum  & 33.5 &  18.7 & 30.9 & 57.6 & 4.3 / 111 \\ & 30.3 &  15.1 &  27.6  & 59.1 & 4.4 / 110 \\ 
     CTRLSum & 32.50 & 17.58 & 29.87 & 58.07 & 4.33 / 110.69 \\
    %  GSUM$_{ovr}$  &  41.1 & 25.8 & 38.2 & 62.1 & 3.6 / 74  \\ & 38.5 & 22.8 & 35.3 & 65.0 & 3.6 / 75 \\
     GSum$_{ovr}$ & 40.29 & 24.87 & 37.3 & 63.00 & 3.60 / 74.31 \\
    %  GSUM$_{ent}$  w/ name &  53.2 & 42.6 & 50.4 & 69.3 & 3.6 / 111  \\ &  48.4 & 35.8 & 45.1 & 71.9 & 3.7 / 111 \\
     GSum$_{ent-name}$ (Ours) & 51.71 & 40.49 & 48.75 & 70.11 & 3.63 / 111.0 \\
    %  GSUM$_{ent}$  w/ sents. &  63.6 & 54.9 & 60.7 & 75.0 & 3.3 / 99 \\ &  56.7 & 45.7 & 53.2 & 77.8 & 3.4 / 101 \\
     GSum$_{ent-sent}$ (Ours) & \textbf{61.45} & \textbf{52.04} & \textbf{58.37} & \textbf{75.87} & 3.33 / 99.62 \\
    \hline
    \rowcolor{lightGray}    \multicolumn{6}{l}{\textbf{Methods using Oracle Entity Sentence Information}} \\
    \hline
    % Lead3$_{ent}$ (salient)  & 79.4 & 74.1 & 76.4 & 84.8 & 2.7 / 91  \\ &  67.4 & 58.6 & 63.5 & 85.9 & 2.8 / 92 \\
    Lead3$_{ent}$ (Salient)  & 75.67 & 69.28 & 72.39 & 85.14 & 2.73 / 91.31 \\
    % Lead3$_{ent}$ (summary) & 91.0 & 87.8 &  88.4 & 91.6 & 2.5 / 86  \\ & 72.4 & 64.3 & 68.5 & 91.2 & 2.6 / 86 \\
    Lead3$_{ent}$ (Summary)  & 85.22 & 80.49 & 82.21 & 91.48 & 2.53 / 86.0 \\
    \hline 
    \end{tabular}
  \caption{Automatic evaluation results of different summarization models on the \textsc{EntSUM} data set. \textbf{Bold} typeface denotes the best performance within a class of methods.}
  \label{table:final_results}
\end{table*}

We automatically evaluate the quality of the generated summaries using unigram and bigram overlap (ROUGE-1 and ROUGE-2), which are a proxy for assessing informativeness and use the longest common subsequence (ROUGE-L) to measure fluency \cite{lin-hovy-2003-automatic}. We also use BERTScore \cite{bertscore_zhang} to compute a similarity score for each token in the generated summaries with each token in the reference summaries using contextualized word embeddings provided by BERT \cite{devlin-etal-2019-bert}. BERTScore incorporates semantic information behind sentences, thus can provide better evaluations for cases where ROUGE score fails to account for meaning-preserving lexical and semantic diversity. BERTScore showed to have better correlations with human judgments for natural language generation \cite{bertscore_zhang}. For the samples in \textsc{EntSUM} where we have multiple reference summaries, we take the maximum ROUGE or BERTScore scores. We also report the average sentence and word lengths of the generated summaries to observe summary statistics for the behavior of the output, as automated metrics are sensitive to summary length.

\section{Results}

We benchmark all methods described above on the newly proposed \textsc{EntSUM} data set in order to establish baseline performance of both abstractive and extractive methods for this new task and data set. Table \ref{table:final_results} shows the automatic evaluation results.

The results show the following trends across all four evaluation metrics:

\textbf{Entity-centric summarization is very different to generic summarization} given that methods that do not take entity information into account (Lead3$_{ovr}$, GSum$_{ovr}$) perform significantly lower than the best methods in the same class which use entity information.

\textbf{Previously introduced methods (ConvNet, CTRLSum) for controllable summarization can not perform well on entity-centric summarization} with their results being over 17 BERTScore and 29 ROUGE-L lower than the proposed adaptation for abstractive summarization on entity-centric summaries. Further, these methods actually obtain lower results by 4.93 BERTScore and 7.43 ROUGE-L than the entity-agnostic GSum$_{ovr}$ method, which shows these methods are not effective at modelling entity-centric information through their training and inference process.

\textbf{Our proposed adaptations to both abstractive and extractive methods perform well} on entity-centric evaluation, despite they were trained on a data set that used proxies for entity-centric summaries. For extractive summarization BERTSum$_{ent-top3}$ performs better than BERTSum$_{ovr}$ by 34.23 ROUGE-L and by 19.65 on BERTScore, while for abstractive summarization GSum$_{ent-sent}$ is better than GSum$_{ovr}$ by 21.07 ROUGE-L and 12.87 BERTScore. We also see that the choice of guidance signal in the GSum framework is impactful, with using sentences with entities leading to 9.62 ROUGE-L and 5.76 BERTScore improvements over using the entity name.

\textbf{Extractive approaches perform better than abstractive} methods, which is expected due to the extractive nature of the \textsc{EntSUM} data set, the gap between the best performing methods (BERTSum$_{ent-top3}$ and GSum$_{ent-sent}$) is clear, when using BERTScore (+2.02) which better estimates semantic similarity opposed to the n-gram matches used in ROUGE (+7.66 on ROUGE-2, +6.03 on ROUGE-L).

\textbf{Lead3$_{ent}$ is a very strong baseline} as expected, because this is a strong baseline for document summarization in general and especially because \textsc{EntSUM} is by design a more extractive summarization data set.

\textbf{Lead3 using oracle selected sentences perform much better than Lead3} and shows the benefits of selecting salient sentences (+7.36 ROUGE-L, +5.16 BERTScore) and the benefits of selecting the most important sentences used in writing the summary (further +9.82 ROUGE-L, +6.26 BERTScore compared to top salient sentences).

The absolute results also show there is \textbf{further room for improvement in entity-centric summarization} approaches, given that performance of automated methods still lags behind Lead3$_{ent}$, whereas this is currently surpassed by automated methods in generic summarization. 

\section{Conclusion}

We introduced the first annotated data set (\textsc{EntSUM}) for controllable summarization where entities are targets for control. We conducted a quantitative analysis of the newly created resource and highlighted how this is different to generic summarization methods. We used the \textsc{EntSUM} data set for benchmarking state-of-the-art generic abstractive and extractive summarization methods, as well as initial methods for controllable summarization. Further, we proposed a new setup for learning entity-centric summaries from generic summarization data sets and, extending previous methods, demonstrated good performance on the newly proposed \textsc{EntSUM} data set.

In the future, we aim to propose new methods for both extractive and abstractive summarization performance through modelling information about the document and the entity in a more complex way. We also plan to create a data set for entity-centric summarization that is more abstractive in nature.

\section*{Acknowledgements}

We would like to thank Chen-Tse Tsai, Umut Topkara and the other members of the NLP team and the broader Bloomberg AI group, who provided invaluable feedback on the task framing and experiments. We wish to thank Wei Xu for supporting the collaboration. We are grateful to our annotators for their diligence in performing this annotation task. 

% Entries for the entire Anthology, followed by custom entries
\bibliography{anthology,custom}
\bibliographystyle{acl_natbib}

\end{document}